\pdfoutput=1

\documentclass[11pt]{article}

\usepackage[]{ACL2023}

\usepackage{times}
\usepackage{latexsym}

\usepackage[T1]{fontenc}

\usepackage{microtype}

\usepackage{inconsolata}
\usepackage{amsmath}
\usepackage{graphicx}
\usepackage{multirow}
\usepackage{soul}
\usepackage{pifont}
\usepackage{float}
\usepackage{times}
\usepackage{latexsym}
\usepackage{graphicx}
\usepackage{booktabs}
\usepackage{subcaption}
\usepackage{multirow}
\usepackage{bbold}
\usepackage{mathtools}
\usepackage{xcolor}
\usepackage{booktabs}
\usepackage{tabularx,ragged2e}
\usepackage{enumitem}
\usepackage{xcolor}
\usepackage{algorithm}
\usepackage{algorithmic}
\usepackage{amsmath}
\usepackage{amssymb}

\definecolor{darkgreen}{rgb}{0,0.5,0}

\DeclareMathOperator*{\argmax}{argmax}

\title{Embarrassingly Simple Unsupervised Aspect Based Sentiment Tuple Extraction}

\author
{Kevin Scaria$^{1*}$ 
\quad 
Abyn Scaria$^{2\spadesuit}$ \quad 
Ben Scaria$^{\dagger}$\quad
\\
\small{$^{1}$Arizona State University} \quad
\small{$^{2}$Georgia Institute of Technology}
\\ \tt\small {\{kscaria\}}@asu.edu
}

\begin{document}
\maketitle
\begin{abstract}
Aspect Based Sentiment Analysis (ABSA) tasks involve the extraction of fine-grained sentiment tuples from sentences, aiming to discern the author's opinions. 
Conventional methodologies predominantly rely on supervised approaches; however, the efficacy of such methods diminishes in low-resource domains lacking labeled datasets since they often lack the ability to generalize across domains.
To address this challenge, we propose a simple and novel unsupervised approach to extract opinion terms and the corresponding sentiment polarity for aspect terms in a sentence. 
Our experimental evaluations, conducted on four benchmark datasets, demonstrate compelling performance to extract the aspect oriented opinion words as well as assigning sentiment polarity.
Additionally, unsupervised approaches for opinion word mining have not been explored and our work establishes a benchmark for the same
\footnote{Experiments and results are available at 
\url{https://anonymous.4open.science/r/AutoABSA}
}.
\end{abstract}

\section{Introduction}
Existing approaches for aspect based sentiment analysis (ABSA) subtasks predominantly fall into two categories: supervised and unsupervised methodologies. 
Recent trends in supervised approaches have seen the formulation of aspect term sentiment classification tasks as an entailment problem \cite{seoh-etal-2021-open}, machine reading comprehension (MRC) problem \cite{DBLP:conf/aaai/ChenWLW21-BMRC}, grid-tagging and table-filling schemes \cite{wu-etal-2020-grid, zhai-etal-2023-ussa}, leveraging techniques such as graph convolution networks \cite{li-etal-2021-dual-graph, chen-etal-2022-enhanced, liang-etal-2022-bisyn}, generative approaches \cite{yan-etal-2021-unified, zhang-etal-2021-towards-generative}, prompt tuning \cite{li2021sentiprompt}, instruction tuning \cite{varia-etal-2023-instruction, scaria2023instructabsa}, multi-task learning \cite{wang2022unifiedabsa}, and supervised contrastive learning \cite{DBLP:journals/corr/abs-2211-07743}, which inherently rely on human-labelled annotated datasets. 

\begin{figure}[t!]
	\includegraphics[width= \linewidth, height= 4.4 cm]{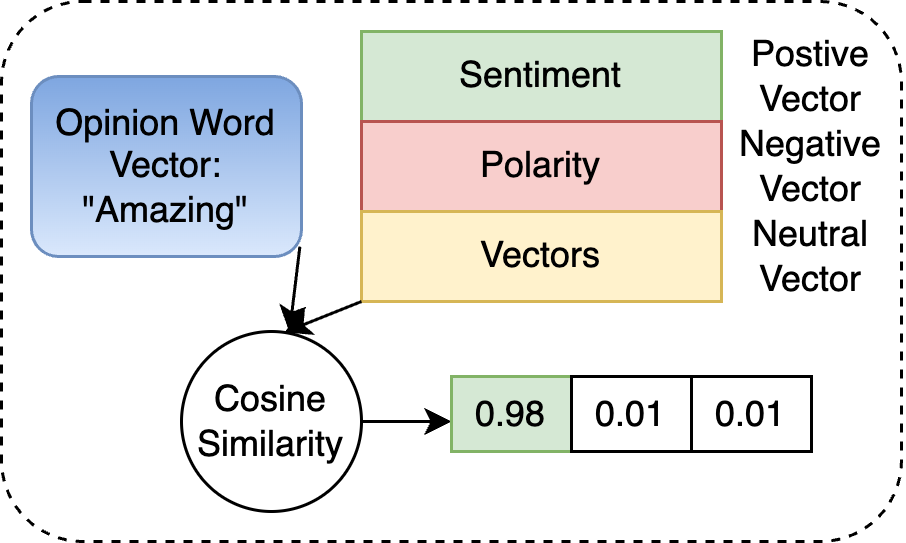}
	\caption{Approach overview for ATSC subtask}
	\label{fig:atsc_teaser}
\end{figure} 
\noindent Furthermore, approaches employing large language models (LLMs) also face challenges in adapting to new domains and necessitate labeled data for fine-tuning. 

\begin{figure*}[t!]
	\centering
	\includegraphics[
 width=\linewidth , 
 height= 5.5  cm]{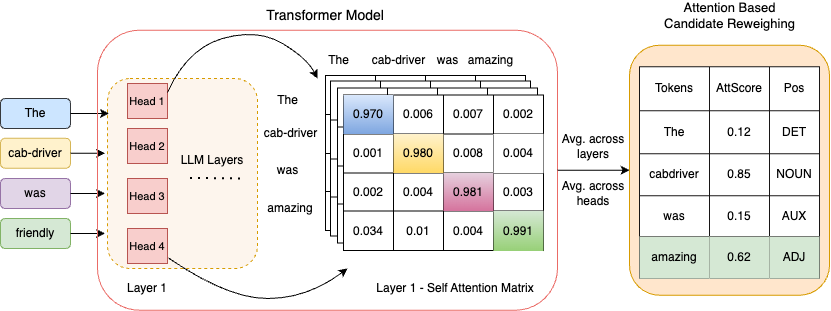}
	\caption{An overview of our approach for AOOE subtask}
	\label{fig:flowchart}
\end{figure*} 

In practice, the curation of annotated datasets proves to be costly and time-consuming. 
Numerous unsupervised methodologies have been proposed for aspect term extraction, employing techniques such as Latent Dirichlet Allocation (LDA) \cite{8923417, VENUGOPALAN2022108668}, topic modeling \cite{PATHAN2021492}, and hybrid candidate filtering \cite{mai-zhang-2020-aspect}. 
Similarly, for aspect category extraction, several unsupervised approaches have been introduced, leveraging contrastive attention \cite{tulkens-van-cranenburgh-2020-embarrassingly} and attention-based aspect extraction \cite{he-etal-2017-unsupervised}. 
However, there have been very limited unsupervised approaches proposed for other complex ABSA subtasks such as aspect oriented opinion (AOOE) extraction and aspect term sentiment classification (ATSC).

We present a simple unsupervised method for the AOOE and ATSC subtasks that only necessitates a Part-of-Speech (POS) tagger and domain-adapted word embeddings trained on the relevant dataset. 
To the best of our knowledge, our work is the first to establish a benchmark for the aforementioned ABSA subtasks in an unsupervised setting. We carry out extensive experiments on several datasets and the results demonstrate the effectiveness of our approach in extracting meaningful opinion words and assigning sentiment polarity to aspects. 
 Additionally, availability of labelled instances improves the performance by $\sim 1.8\%$ across subtasks.  
 Notably, further analysis reveals that scaling up the model size improves the performance by $\sim 2\%$ across subtasks. 
 Furthermore, our findings indicate evidence of domain generalizability.

\section{Task Defintion}

Consider the $i^{th}$ review sentence, denoted as $S_i$, in the training sample, where $S_i = \{w_{i}^1, w_{i}^2, ..., w_{i}^n\}$ and $n$ is the number of tokens in the sentence. Each $S_i$ contains a set of aspect terms, represented by $A_i = \{a_{i}^1, a_{i}^2, ..., a_{i}^m\}$ with $m \le n$. The corresponding opinion terms and sentiment polarities for each aspect term are denoted by $O_{i} = \{o_{i}^1, o_{i}^2, ..., o_{i}^m\}$ and $SP_{i} = \{sp_{i}^1, sp_{i}^2, ..., sp_{i}^m\}$, where $sp_i^k \in \{ positive, negative, neutral \}$.

The system takes both the sentence $S_i$ and the aspect term $a_{i}^k$ as the input. For the AOPE subtask, the opinion word $o_{i}^k$ corresponding to the aspect term is extracted whereas for the ATSC subtask, the corresponding sentiment polarity $sp_i^k$ is assigned. The Aspect Oriented Opinion Sentiment Pair Extraction (AOOSPE) subtask aims to extract the opinion-sentiment tuple $(o_{i}^k, sp_i^k)$ for each aspect term in a given sentence. 

\section{Method}

\subsection{Domain Adaptation}

In our proposed method, the first step is domain adaptation. 
To acquire in-domain embeddings, we conduct pre-finetuning of a language model \(LM\) using an in-domain dataset. 






\subsection{Compound Phrase Extraction}

Previous works \cite{10.1145/1014052.1014073, mai-zhang-2020-aspect, tulkens-van-cranenburgh-2020-embarrassingly} use POS taggers and dependency trees to extract NOUN chunks for extracting aspect terms.
However, in the second step of our pipeline, we extract opinion words by identifying compound phrases. 
This involves extracting nouns that are modified by sentiment-bearing adjectives. This opinion relation enables the association of an opinion word with the aspect term, and subsequently in the assignment of sentiment polarity.
For each $S_i$, we investigate the presence of five opinion relations: "Adj-\{mod\}-Noun", "Adv-\{mod\}-Adj", "Adj-\{mod\}-Noun-\{mod\}-Noun", and "Adp-\{mod\}-Noun".

\subsection{Attention Weighting}
\label{sec:att_weighting}

The opinion words extracted using the opinion relations have very poor precision \cite{xu-etal-2013-mining}, owing to false opinion relations, false opinion targets, and scenarios where there are long-tail opinion targets. 
It becomes challenging to associate a candidate opinion word with an aspect term and simultaneously filter out non-opinion words. 
Thus as part of this step, we weigh the candidate opinion terms using the self-attention scores corresponding to the aspect term.
The self-attention matrix for the review sentence $S_i$ with $n$ tokens from a given layer $N_\text{layer}$ of the model $LM$ is obtained using the scaled dot-product attention.
\[AttScore_{[a_i^k]} = \text{softmax}\left(\frac{Q_{[a_i^k]} \cdot K^T}{\sqrt{d_k}}\right) \]

Where $a_i^k$ is the $k^{th}$ aspect term and $Q$, $K$ is the query and key matrices of the $n$ tokens for the review sentence $S_i$. 
\(d_k\) is the dimensionality of the key matrix. 
The candidate opinion word corresponding to the index of the maximum attention score w.r.t the aspect term is finally selected as the opinion word.
\[o_i^k  = \text{argmax} \left(AttScore_{[a_i^k]} \cdot I \right) 
\]
The indicator function \(I\) is defined as follows:
\[I(w_i^k) =
\begin{cases}
  1, & \text{if } w_i^k \text{ is a candidate opinion word} \\
  0, & \text{otherwise}
\end{cases}
\]

Since transformer models, encode the linguistic syntactic structure in the lower layers \cite{jawahar-etal-2019-bert}, we use attention scores from the first four layers of the $LM$ model. 
For each $N_\text{layer}$, we aggregate the attention scores across the $N_\text{heads}$ attention heads.
\[o_i^k  = \text{argmax} \left(\frac{1}{N_{\text{heads}}} \sum_{N_\text{heads}} \left(AttScore_{[a_i^k]}  \right) \cdot I \right) 
\]

\subsection{Semantic Polarity Assignment}
In this final step, we generate representations $h_{op}$ using the model $LM$ for the opinion terms extracted in the previous step. 
To assign the semantic polarity to each opinion term, we compute the cosine similarity between $h_o$ and the polarity label vectors.
\[sp_i^k = \argmax_{c \in C} \left(cos(h_{op} \cdot \vec{c}) \right)\]

Where, $C$ is the set of labels used to assign the sentiment polarity labels i.e. $\{positive, negative, neutral\}$.

\section{Results and Analysis}
\label{sec_results}

\subsection{Experimental Setup}
 
 We utilize the SemEval 2014, 2015, and 2016 datasets for our experimentation. 
These datasets serve as benchmarks for ABSA tasks and comprise customer reviews from three domains: laptops (L14), hotels (H15), and restaurants (R14, R15, and R16).
Following previous work we report the performance only on the laptops and restaurants dataset.
All the results reported are the average values from 5 runs of the domain adaptation step for each experiment.
Further details and results regarding experiments on other MLMs and auto-regressive models are available in \S \ref{sec:experimental_setup}, \ref{sec:dataset} \& \ref{sec:further_results}.

\subsection{Sub Task Results} 

Table \ref{tab:aooe_result}, \ref{tab:atsc_result}, and \ref{tab:aoospe_result} denotes the results of AOOE, ATSC, and AOOSPE subtasks respectively. 

\begin{table}[H]
\setlength\tabcolsep{4.0pt}
\setlength{\belowcaptionskip}{-10pt}
\centering
\scriptsize
\resizebox{\linewidth}{!}{
    \begin{tabular}{l|r|r|r|r}
\toprule
 {\textbf{Model}} & \multicolumn{1}{l|}{\textbf{L14}} & \multicolumn{1}{l|}{\textbf{R14}} & \multicolumn{1}{l|}{\textbf{R15}} & \multicolumn{1}{l}{\textbf{R16}} \\ \midrule
ELECTRA & \textbf{51.98} & \textbf{62.84} & \textbf{65.32} & \textbf{66.67} \\
DeBERTa  & 50.18 & 58.41 & 62.32 & 65.11\\
T5  & 39.98 & 40.18 & 45.65 & 46.67\\
Pythia & 47.76 & 60.12 & 61.32 & 60.21\\
\bottomrule
\end{tabular}
}
\caption{Results of the AOOE reporting the accuracy.}
\label{tab:aooe_result}
\end{table}

\noindent For the \textbf{AOOE} subtask (refer to Table \ref{tab:aooe_result}), the ELECTRA model has consistently achieved the highest scores across all four datasets, outperforming other models. 
Following ELECTRA, Pythia, a decoder-only model, has also shown notable performance, while the seq2seq T5 model has consistently yielded lower scores. Furthermore, encoder-only models exhibit  $\sim 3.58\%$ better performance compared to decoder-only models. 
This trend can be attributed to the encoder models' capability to better capture the contextual information necessary for extracting opinion words."

\begin{table}[H]
\setlength\tabcolsep{4.0pt}
\setlength{\belowcaptionskip}{-10pt}
\centering
\scriptsize
\resizebox{\linewidth}{!}{
    \begin{tabular}{l|r|r|r|r}
\toprule
 {\textbf{Model}} & \multicolumn{1}{l|}{\textbf{L14}} & \multicolumn{1}{l|}{\textbf{R14}} & \multicolumn{1}{l|}{\textbf{R15}} & \multicolumn{1}{l}{\textbf{R16}} \\ \midrule
ELECTRA & \textbf{54.34} & \textbf{62.13} & \textbf{61.78} & 61.93 \\ 
DeBERTa  & 52.21 & 60.34 & 61.43 & \textbf{63.23}\\
T5  & 53.69 & 57.30 & 56.61 & 57.22\\
Pythia & 52.26 & 60.13 & 61.21 & 61.84\\
\bottomrule
\end{tabular}
}
\caption{Results of the ATSC reporting the accuracy.}
\label{tab:atsc_result}
\end{table}

\noindent In the \textbf{ATSC} subtask (refer to Table \ref{tab:atsc_result}), the ELECTRA model consistently outperforms other models on the L14, R14, and R15 datasets. However, on the R16 dataset, the DeBERTa model achieved the highest accuracy score. 

\begin{table}[H]
\setlength\tabcolsep{4.0pt}
\setlength{\belowcaptionskip}{-10pt}
\centering
\scriptsize
\resizebox{\linewidth}{!}{
    \begin{tabular}{l|r|r|r|r}
\toprule
 {\textbf{Model}} & \multicolumn{1}{l|}{\textbf{L14}} & \multicolumn{1}{l|}{\textbf{R14}} & \multicolumn{1}{l|}{\textbf{R15}} & \multicolumn{1}{l}{\textbf{R16}} \\ \midrule
ELECTRA & \textbf{51.76} & \textbf{61.39} & \textbf{61.25} & 60.95 \\
DeBERTa  & 49.56 & 57.12 & 60.34 & \textbf{62.77}\\
T5  & 34.54 & 39.31 & 43.69 & 44.17\\
Pythia & 45.12 & 58.33 & 60.17 & 59.25\\
\bottomrule
\end{tabular}
}
\caption{Results of the AOOSPE reporting the accuracy.}
\label{tab:aoospe_result}
\end{table}

\noindent Similarly for the \textbf{AOOSPE} subtask (Table \ref{tab:aoospe_result}), the ELECTRA model performs best for the L14, R14 and R15 dataset, whereas for the R16 dataset, DeBERTa outperforms other models.

\begin{table*}[h]
\resizebox{\linewidth}{!}{
\begin{tabular}{l|rrrrrr|rrrrrr}
\hline
\multicolumn{1}{c|}{\multirow{3}{*}{Model}} & \multicolumn{6}{c|}{Cross Domain}                                                                                                                                         & \multicolumn{6}{c}{Join Domain}                                                                                                                                          \\ \cline{2-13} 
\multicolumn{1}{c|}{}                       & \multicolumn{2}{l|}{AOOE}                               & \multicolumn{2}{l|}{ATSC}                               & \multicolumn{2}{l|}{AOOSTE}                           & \multicolumn{2}{l|}{AOOE}                               & \multicolumn{2}{l|}{ATSC}                               & \multicolumn{2}{l}{AOOSTE}                           \\ \cline{2-13} 
\multicolumn{1}{c|}{}                       & \multicolumn{1}{l|}{L14}   & \multicolumn{1}{l|}{R14}   & \multicolumn{1}{l|}{L14}   & \multicolumn{1}{l|}{R14}   & \multicolumn{1}{l|}{L14}   & \multicolumn{1}{l|}{R14} & \multicolumn{1}{l|}{L14}   & \multicolumn{1}{l|}{R14}   & \multicolumn{1}{l|}{L14}   & \multicolumn{1}{l|}{R14}   & \multicolumn{1}{l|}{L14}   & \multicolumn{1}{l}{R14} \\ \hline
ELECTRA                                    & \multicolumn{1}{r|}{50.79} & \multicolumn{1}{r|}{59.11} & \multicolumn{1}{r|}{49.99} & \multicolumn{1}{r|}{59.62} & \multicolumn{1}{r|}{49.70} & 56.76                    & \multicolumn{1}{r|}{53.43} & \multicolumn{1}{r|}{63.79} & \multicolumn{1}{r|}{55.79} & \multicolumn{1}{r|}{63.32} & \multicolumn{1}{r|}{52.96} & 62.37                   \\
DeBERTa                                     & \multicolumn{1}{r|}{48.98}     & \multicolumn{1}{r|}{54.78}     & \multicolumn{1}{r|}{47.59}     & \multicolumn{1}{r|}{59.43}     & \multicolumn{1}{r|}{47.88}     & 54.73                        & \multicolumn{1}{r|}{50.95}     & \multicolumn{1}{r|}{58.71}     & \multicolumn{1}{r|}{52.54}     & \multicolumn{1}{r|}{61.05}     & \multicolumn{1}{r|}{50.21}     & 58.11                       \\
T5                                          & \multicolumn{1}{r|}{38.79}     & \multicolumn{1}{r|}{38.09}     & \multicolumn{1}{r|}{49.24}     & \multicolumn{1}{r|}{51.42}     & \multicolumn{1}{r|}{32.86}     & 33.54                        & \multicolumn{1}{r|}{41.01}     & \multicolumn{1}{r|}{40.65}     & \multicolumn{1}{r|}{53.87}     & \multicolumn{1}{r|}{57.98}     & \multicolumn{1}{r|}{35.02}     & 40.43                       \\
Pythia                                      & \multicolumn{1}{r|}{46.82}     & \multicolumn{1}{r|}{55.99}     & \multicolumn{1}{r|}{47.89}     & \multicolumn{1}{r|}{56.26}     & \multicolumn{1}{r|}{44.23}     & 55.23                        & \multicolumn{1}{r|}{48.31}     & \multicolumn{1}{r|}{61.11}     & \multicolumn{1}{r|}{52.43}     & \multicolumn{1}{r|}{60.93}     & \multicolumn{1}{r|}{46.55}     & 59.65                       \\
\cline{1-13}
\end{tabular}
}
\caption{Results of joint and cross domain adaptation results.}
\label{tab:main_joint_cross_domain}
\end{table*}

Across all three subtask, the ELECTRA model consistently yields higher performance. 
This could be due to better contextual representations, acquired through its replaced token detection objective, as compared to those of other encoder models. 
By learning from the entire input sequence rather than just the masked segments, ELECTRA exhibits significant benefits in the AOOE subtask, which leverages the linguistic structural information encoded in lower layers for extracting opinion terms. 

\subsection{Analysis}
\label{analysis}

In this subsection, we analyze our approach on multiple line of enquiries and carry out ablation studies in \S\ref{sec: ablation_studies}.

\begin{figure}[h!]
	\centering
	\includegraphics[
 width=\linewidth , 
 height= 6.5  cm]{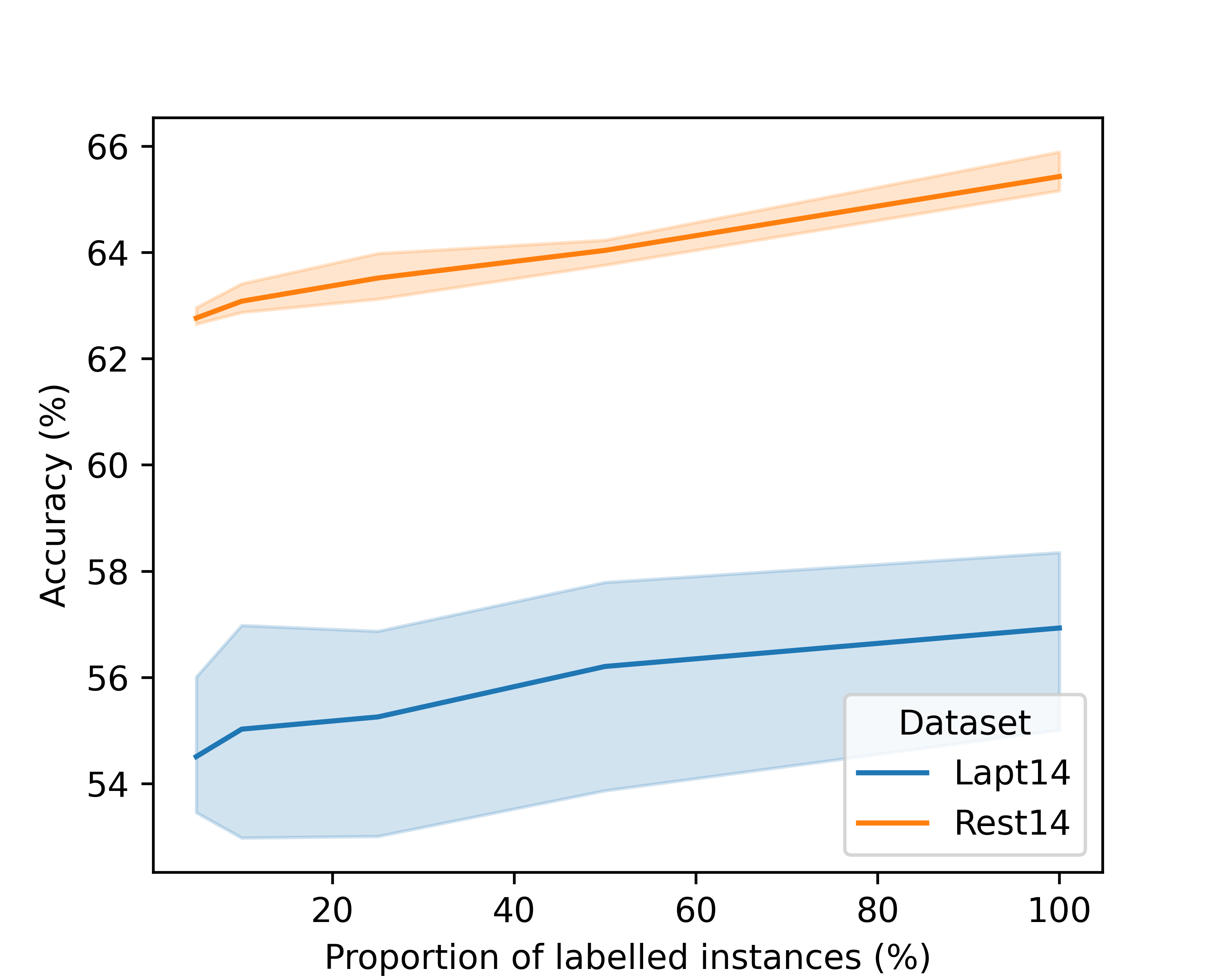}
	\caption{Accuracy across tasks when different \% of labelled instances are available by the ELECTRA model.}
	\label{fig:labelled_data_availability}
\end{figure} 

\paragraph{Availability of labelled instances improves performance: }
We inspect the performance of our approach when labelles instances are available. We model ATSC as a sequence classification task using the best performing model ELECTRA. We carry out the following scenarios where there are 5\%, 10\%, 25\%, 50\%, and 100\% of labelled instance availability of sentiment polarity labels of the training set. From figure \ref{fig:labelled_data_availability}, it can be seen that upon using the finetuned models, across all tasks, there is an improvement of $\sim 2.8\%$ on average. The availability of polarity labels enhances the attention matrix which helps in better improving the extraction of aspect terms.

\paragraph{Competitive cross-domain generalization: } 
In cross domain setting, we perform the domain adaptation step on a train set from one domain and evaluate on a test set from different domain. 
From table \ref{tab:main_joint_cross_domain} it can be seen that our approach has comparable ($\sim 90\%$) performance to in-domain adaptation. Additionally, the cross-domain performance is $\sim 1.1$ higher for encoder-only models as compared to decoder-only models.

\paragraph{Joint-domain adaptation improves performance:} 
In joint domain setting, the train data of the domains (laptops and restaurants) are combined to perform the domain adaptation step, and evaluation is done on individual test sets. From table \ref{tab:main_joint_cross_domain} it is evident that using joint domain adaptation the performance of our approach increases by $\sim 1.9\%$. Notably, the joint domain adaptation improves the performance of the AOOSPE task the most. Similar to previous trend, the encoder-only models achieve higher performance gains ($\sim 0.4\%$ points) as compared to decoder-only models.

\section{Conclusion}
We proposed a novel unsupervised learning approach to address the two ABSA subtasks viz. ATSC and AOOE. 
Our results establish a strong benchmark for these subtasks in an unsupervised setting but also demonstrate competitive performance compared to various supervised approaches. 
Our approach does not require the presence of labeled datasets, showcasing its adaptability to low-resource domains. 
Finally we have made our code publicly available. We hope that our contributions will stimulate further exploration and advancements in unsupervised ABSA methodologies.

\section*{Limitations}
Our study confines itself to the widely employed SemEval 2014, 15, and 16 datasets, representing a common practice in recent research. 
To ensure the broader applicability of our findings, future investigations should expand this work to encompass other ABSA datasets, thereby validating the generalizability of our approach. 
It is noteworthy that our experiments exclusively utilized the base variants of the models, each ranging from 100 to 200 million parameters. Subsequent studies might explore the impact of employing even larger models to gain deeper insights into their performance nuances. 
Another limitation lies in our focus on the English language, implying that our conclusions may not seamlessly translate to other languages. 
Addressing this, future research endeavors should encompass a diverse set of languages, incorporating multilingual datasets and other models.

\section*{Ethical Considerations}
We recognize the potential presence of inherent biases in the language models employed, stemming from the characteristics of the pre-training data. Although our study did not specifically conduct stress testing, our comprehensive analysis indicates that no discernible additional concerns are associated with privacy, fairness, bias, or discrimination. 
By focusing on aspect-based sentiment analysis, our research contributes directly to the scientific discourse in a constructive manner. 
We are optimistic about the positive impact of our work on the broader scientific community. 
Our commitment to responsible AI usage remains unwavering, and we pledge to consistently prioritize ethical considerations in all our future research endeavors.

\bibliography{anthology,custom}
\bibliographystyle{acl_natbib}

\clearpage

\section*{Appendix}
\appendix

\section{Ablation Studies}
\label{sec: ablation_studies}
\paragraph{Impact of Domain Adaptation:}
We investigate the impact of the domain adaptation step. The table \ref{tab:ablation_domain_adaptation} presents the performance of the best performing model ELECTRA with and without the domain adaptation step. It is evident that domain adaptation improves the performance across subtasks by $\sim 2\%$ on an average.

\begin{table}[H]
\begin{tabular}{l|r|r|r|r}
\hline
\multicolumn{1}{c|}{\textbf{Task}} & \multicolumn{1}{c|}{\textbf{L14}} & \multicolumn{1}{c|}{\textbf{R14}} & \multicolumn{1}{c|}{\textbf{R15}} & \multicolumn{1}{c}{\textbf{R16}} \\ \hline
AOOE$_{da}$                            & 51.98                             & 62.64                             & 65.32                             & 66.67                            \\
AOOE$_{w/o da}$                        & 49.12                             & 60.32                             & 64.11                             & 64.21                            \\
ATSC$_{da}$                           & 54.34                             & 62.13                             & 61.78                             & 61.93                            \\
ATSC$_{w/o da}$                        & 52.76                             & 61.77                             & 60.22                             & 59.45                            \\
AOOSPE$_{da}$                          & 51.76                             & 61.39                             & 61.25                             & 60.95                            \\
AOOSPE$_{w/o da}$                      & 50.54                             & 58.76                             & 60.21                             & 58.89                            \\ \hline
\end{tabular}
\caption{Results with and without domain adaptation results by the best performing model for each subtask. $da$, $w/oda$ stands for domain adaptation and without $da$ respectively.}
\label{tab:ablation_domain_adaptation}
\end{table}

\paragraph{Impact of Massive Domain Finetuning:}
The yelp dataset \cite{asghar2016yelp} is a massive dataset of yelp reviews on restaurants and products. We utilize this dataset as part of the domain adaptation step to learn the rich latent dependency between aspects and opinion terms. The table \ref{tab:ablation_massive_domain_adaptation} presents the performance of the best performing model ELECTRA with and without the massive domain finetuning step. Notably, this improves the performance across subtasks by $\sim 4.2\%$ on an average.

\begin{table}[H]
\begin{tabular}{l|r|r|r|r}
\hline
\multicolumn{1}{c|}{\textbf{Task}} & \multicolumn{1}{c|}{\textbf{L14}} & \multicolumn{1}{c|}{\textbf{R14}} & \multicolumn{1}{c|}{\textbf{R15}} & \multicolumn{1}{c}{\textbf{R16}} \\ \hline
AOOE$_{mdf}$                            & 54.78                             & 65.76                             & 67.43                             & 68.12                            \\
ATSC$_{mdf}$                           & 57.32                             & 64.54                             & 63.34                            & 63.37                            \\
AOOSPE$_{mdf}$                          & 59.01                             & 64.34                             & 64.49                             & 64.81                            \\
\hline
\end{tabular}
\caption{Results of massive domain finetuning results across subtask by best performing model. $mdf$ stands for massive domain finetuning.}
\label{tab:ablation_massive_domain_adaptation}
\end{table}



\section{Experimental Setup}
\label{sec:experimental_setup}

\paragraph{Models Used:} We experiment with various types of models such as BERT \cite{devlin-etal-2019-bert}, ALBERT \cite{DBLP:conf/iclr/LanCGGSS20}, RoBERTa \cite{zhuang-etal-2021-robustly}, ELECTRA \cite{DBLP:journals/corr/abs-2003-10555} and DeBERTa \cite{DBLP:journals/corr/abs-2006-03654}, T5 \cite{roberts2019exploring}, Pythia \cite{biderman2023pythia}, GPTNeo, and Cerebras \cite{dey2023cerebras}. 
 To ensure a fair comparison of the model's performance, we use the base variant of the MLMs and T5, the 160M parameter model for Pythia, the 125M parameter model for GPT-Neo, and the 111M parameter model for Cerebras.

\paragraph{Hyperparameters}GPU: 1xNvidia Tesla P40,
Domain Adaptation Step Batch Size: 16,
Gradient Accumulation Steps: 2,
Initial learning rate: 5e-5,
Num of Epochs: 5

\paragraph{Evaluation Metric:} We report the accuracy scores for the AOOE, ATSC \& AOOSPE subtasks. 

\section{Further Results}
\label{sec:further_results}

The following tables \ref{tab:apdx_aooe_result}, \ref{tab:apdx_atsc_result} and \ref{tab:apdx_aoospe_result}, present the experiements run on a variety of models as mentioned in \S \ref{sec:experimental_setup}.

\begin{table}[ht!]
\setlength\tabcolsep{4.0pt}
\setlength{\belowcaptionskip}{-10pt}
\centering
\scriptsize
\resizebox{\linewidth}{!}{
    \begin{tabular}{l|r|r|r|r}
\toprule
 {\textbf{Model}} & \multicolumn{1}{l|}{\textbf{L14}} & \multicolumn{1}{l|}{\textbf{R14}} & \multicolumn{1}{l|}{\textbf{R15}} & \multicolumn{1}{l}{\textbf{R16}} \\ \midrule
BERT & 49.79 & 59.01 & 61.98 & 63.74 \\
ALBERT  & 50.37 & 62.54 & 64.26 & 66.25 \\
RoBERTa & 48.08 & 58.81 & 60.33 & 60.82 \\
ELECTRA & \textbf{51.98} & \textbf{62.84} & \textbf{65.32} & \textbf{66.67} \\
DeBERTa  & 50.18 & 58.41 & 62.32 & 65.11\\
T5  & 39.98 & 40.18 & 45.65 & 46.67\\
GPTNeo  & 46.91 & 58.67 & 60.09 & 60.14\\
Pythia & 47.76 & 60.12 & 61.32 & 60.21\\
Cerebras & 49.21 & 56.64 & 60.92 & 61.75\\
\bottomrule
\end{tabular}
}
\caption{Results of the AOOE subtask reporting accuracy scores.}
\label{tab:apdx_aooe_result}
\end{table}

\begin{table}[ht!]
\setlength\tabcolsep{4.0pt}
\setlength{\belowcaptionskip}{-10pt}
\centering
\scriptsize
\resizebox{\linewidth}{!}{
    \begin{tabular}{l|r|r|r|r}
\toprule
 {\textbf{Model}} & \multicolumn{1}{l|}{\textbf{L14}} & \multicolumn{1}{l|}{\textbf{R14}} & \multicolumn{1}{l|}{\textbf{R15}} & \multicolumn{1}{l}{\textbf{R16}} \\ \midrule
BERT & 52.03 & 60.32 & 60.33 & 60.68 \\
ALBERT  & 52.40 & 61.73 & 61.77 & 62.69 \\
RoBERTa & 52.58 & 60.32 & 60.74 & 60.78 \\
ELECTRA & \textbf{54.34} & \textbf{62.13} & \textbf{61.78} & 61.93 \\ 
DeBERTa  & 52.21 & 60.34 & 61.43 & \textbf{63.23}\\
T5  & 53.69 & 57.30 & 56.61 & 57.22\\
GPTNeo  & 51.23 & 61.77 & 60.91 & 61.25\\
Pythia & 52.26 & 60.13 & 61.21 & 61.84\\
Cerebras  & 53.50 & 59.96 & 61.11 & 62.01\\
\bottomrule
\end{tabular}
}
\caption{Results of the ATSC subtask reporting the accuracy scores.}
\label{tab:apdx_atsc_result}
\end{table}

\begin{table*}[h]
\resizebox{\linewidth}{!}{
\begin{tabular}{l|rrrrrr|rrrrrr}
\hline
\multicolumn{1}{c|}{\multirow{3}{*}{Model}} & \multicolumn{6}{c|}{Cross Domain}                                                                                                                                         & \multicolumn{6}{c}{Join Domain}                                                                                                                                          \\ \cline{2-13} 
\multicolumn{1}{c|}{}                       & \multicolumn{2}{l|}{AOOE}                               & \multicolumn{2}{l|}{ATSC}                               & \multicolumn{2}{l|}{AOOSTE}                           & \multicolumn{2}{l|}{AOOE}                               & \multicolumn{2}{l|}{ATSC}                               & \multicolumn{2}{l}{AOOSTE}                           \\ \cline{2-13} 
\multicolumn{1}{c|}{}                       & \multicolumn{1}{l|}{L14}   & \multicolumn{1}{l|}{R14}   & \multicolumn{1}{l|}{L14}   & \multicolumn{1}{l|}{R14}   & \multicolumn{1}{l|}{L14}   & \multicolumn{1}{l|}{R14} & \multicolumn{1}{l|}{L14}   & \multicolumn{1}{l|}{R14}   & \multicolumn{1}{l|}{L14}   & \multicolumn{1}{l|}{R14}   & \multicolumn{1}{l|}{L14}   & \multicolumn{1}{l}{R14} \\ \hline
BERT                                        & \multicolumn{1}{r|}{48.84} & \multicolumn{1}{r|}{57.87} & \multicolumn{1}{r|}{48.01} & \multicolumn{1}{r|}{58.72} & \multicolumn{1}{r|}{47.59} & 56.32                    & \multicolumn{1}{r|}{50.21} & \multicolumn{1}{r|}{60.03} & \multicolumn{1}{r|}{52.88} & \multicolumn{1}{r|}{60.77} & \multicolumn{1}{r|}{49.43} & 58.37                   \\
ALBERT                                      & \multicolumn{1}{r|}{47.99} & \multicolumn{1}{r|}{58.50} & \multicolumn{1}{r|}{47.57} & \multicolumn{1}{r|}{57.35} & \multicolumn{1}{r|}{46.23} & 55.07                    & \multicolumn{1}{r|}{50.80} & \multicolumn{1}{r|}{62.76} & \multicolumn{1}{r|}{52.80} & \multicolumn{1}{r|}{61.75} & \multicolumn{1}{r|}{49.21} & 61.77                   \\
RoBERTa                                     & \multicolumn{1}{r|}{46.91} & \multicolumn{1}{r|}{56.51} & \multicolumn{1}{r|}{48.13} & \multicolumn{1}{r|}{55.37} & \multicolumn{1}{r|}{44.94} & 54.36                    & \multicolumn{1}{r|}{48.43} & \multicolumn{1}{r|}{59.02} & \multicolumn{1}{r|}{53.11} & \multicolumn{1}{r|}{60.91} & \multicolumn{1}{r|}{47.98} & 58.76                   \\
ELECTRA                                    & \multicolumn{1}{r|}{50.79} & \multicolumn{1}{r|}{59.11} & \multicolumn{1}{r|}{49.99} & \multicolumn{1}{r|}{59.62} & \multicolumn{1}{r|}{49.70} & 56.76                    & \multicolumn{1}{r|}{53.43} & \multicolumn{1}{r|}{63.79} & \multicolumn{1}{r|}{55.79} & \multicolumn{1}{r|}{63.32} & \multicolumn{1}{r|}{52.96} & 62.37                   \\
DeBERTa                                     & \multicolumn{1}{r|}{48.98}     & \multicolumn{1}{r|}{54.78}     & \multicolumn{1}{r|}{47.59}     & \multicolumn{1}{r|}{59.43}     & \multicolumn{1}{r|}{47.88}     & 54.73                        & \multicolumn{1}{r|}{50.95}     & \multicolumn{1}{r|}{58.71}     & \multicolumn{1}{r|}{52.54}     & \multicolumn{1}{r|}{61.05}     & \multicolumn{1}{r|}{50.21}     & 58.11                       \\
T5                                          & \multicolumn{1}{r|}{38.79}     & \multicolumn{1}{r|}{38.09}     & \multicolumn{1}{r|}{49.24}     & \multicolumn{1}{r|}{51.42}     & \multicolumn{1}{r|}{32.86}     & 33.54                        & \multicolumn{1}{r|}{41.01}     & \multicolumn{1}{r|}{40.65}     & \multicolumn{1}{r|}{53.87}     & \multicolumn{1}{r|}{57.98}     & \multicolumn{1}{r|}{35.02}     & 40.43                       \\
GPTNeo                                      & \multicolumn{1}{r|}{48.79}     & \multicolumn{1}{r|}{56.55}     & \multicolumn{1}{r|}{46.38}     & \multicolumn{1}{r|}{58.28}     & \multicolumn{1}{r|}{45.65}     & 56.05                        & \multicolumn{1}{r|}{50.87}     & \multicolumn{1}{r|}{58.93}     & \multicolumn{1}{r|}{52.31}     & \multicolumn{1}{r|}{62.14}     & \multicolumn{1}{r|}{47.93}     & 58.65                       \\
Pythia                                      & \multicolumn{1}{r|}{46.82}     & \multicolumn{1}{r|}{55.99}     & \multicolumn{1}{r|}{47.89}     & \multicolumn{1}{r|}{56.26}     & \multicolumn{1}{r|}{44.23}     & 55.23                        & \multicolumn{1}{r|}{48.31}     & \multicolumn{1}{r|}{61.11}     & \multicolumn{1}{r|}{52.43}     & \multicolumn{1}{r|}{60.93}     & \multicolumn{1}{r|}{46.55}     & 59.65                       \\
Cerebras                                    & \multicolumn{1}{r|}{46.93}     & \multicolumn{1}{r|}{53.65}     & \multicolumn{1}{r|}{48.72}     & \multicolumn{1}{r|}{56.07}     & \multicolumn{1}{r|}{44.56}     & 54.56                        & \multicolumn{1}{r|}{48.81}     & \multicolumn{1}{r|}{57.09}     & \multicolumn{1}{r|}{54.04}     & \multicolumn{1}{r|}{59.98}     & \multicolumn{1}{r|}{46.97}     & 56.21                       \\ \cline{1-13}
\end{tabular}
}
\caption{Results of joint and cross-domain adaptation results.}
\label{tab:apdx_joint_cross_domain}
\end{table*}

\begin{table}[ht!]
\setlength\tabcolsep{4.0pt}
\setlength{\belowcaptionskip}{-10pt}
\centering
\scriptsize
\resizebox{\linewidth}{!}{
    \begin{tabular}{l|r|r|r|r}
\toprule
 {\textbf{Model}} & \multicolumn{1}{l|}{\textbf{L14}} & \multicolumn{1}{l|}{\textbf{R14}} & \multicolumn{1}{l|}{\textbf{R15}} & \multicolumn{1}{l}{\textbf{R16}} \\ \midrule
BERT & 48.12 & 58.23 & 60.02 & 59.91 \\
ALBERT  & 48.35 & 61.20 & 60.85 & 61.76 \\
RoBERTa & 47.69 & 58.21 & 59.87 & 58.65 \\
ELECTRA & 51.76 & 61.39 & 61.25 & 60.95 \\
DeBERTa  & 49.56 & 57.12 & 60.34 & 62.77\\
T5  & 34.54 & 39.31 & 43.69 & 44.17\\
Pythia  & 47.83 & 57.87 & 58.21 & 60.06\\
GPTNeo  & 45.12 & 58.33 & 60.17 & 59.25\\
Cerebras  & 46.53 & 55.96 & 59.88 & 57.53\\
\bottomrule
\end{tabular}
}
\caption{Results of the AOOSPE reporting the accuracy scores.}
\label{tab:apdx_aoospe_result}
\end{table}

\section{Extended Related Work}

LMs and deep learning methods have been used for a plethora of downstream tasks for a long time.
Several recent works have leveraged NLP methods and simple sampling methods for different downstream results 
The study of whether existing LMs can understand instructions has motivated a range of subsequent works. 
\citet{gupta2022john} showed the feasibility of LLM's understanding of context. This was leveraged extensively for creating the attention maps for opinion word extraction as well as sentiment polarity assignment.
Furthermore, recent approaches have been proposed to improve the use of targeted data generation for synthetic data creation which could be used as a cost-effective alternative for data labelling \cite{gupta2023targen}. 

\end{document}